\documentclass[pdflatex,sn-mathphys-num]{sn-jnl}

\usepackage{graphicx}
\usepackage{multirow}
\usepackage{amsmath,amssymb,amsfonts}
\usepackage{amsthm}
\usepackage{mathrsfs}
\usepackage[title]{appendix}
\usepackage{xcolor}
\usepackage{textcomp}
\usepackage{manyfoot}
\usepackage{booktabs}
\usepackage{algorithm}
\usepackage{algorithmicx}
\usepackage{algpseudocode}
\usepackage{listings}
\usepackage{csquotes}
\usepackage{subfig}
\usepackage{tabularx}

\theoremstyle{thmstyleone}%
\theoremstyle{thmstyletwo}%

\theoremstyle{thmstylethree}%

\begin{document}

\title[HeartBeatAI]{HeartBeatAI: An Interpretable and Robust Deep Learning Framework for Multi-Label ECG Arrhythmia Detection}

\author*[1]{\fnm{Shubham} \sur{Gupta}}\email{24et0018@iitism.ac.in}
\author[2]{\fnm{Nikhil} \sur{Panwar}}\email{nikhil\_p@cs.iitr.ac.in}
\author[1]{\fnm{Partha Pratim} \sur{Roy}}\email{parthapratim@iitism.ac.in}

\affil*[1]{\orgdiv{Department of Computer Science and Engineering}, \orgname{Indian Institute of Technology (ISM) Dhanbad}, \city{Dhanbad}, \state{Jharkhand}, \country{India}}

\affil[2]{\orgdiv{Department of Computer Science and Engineering}, \orgname{Indian Institute of Technology Roorkee}, \city{Roorkee}, \state{Uttarakhand}, \country{India}}









\abstract{While Deep Learning (DL) enhances automated electrocardiogram (ECG) analysis, clinical deployment is hindered by class imbalance and the generalization gap. This paper presents HeartBeatAI, a deep learning framework combining domain generalization, multi-scale feature aggregation, and clinical explainability for robust 12-lead ECG classification. Moving beyond image-based paradigms, HeartBeatAI integrates a Squeeze-and-Excitation (SE) ResNet to isolate diagnostic leads alongside a Multi-Layer Concentration Pipeline to capture macro-rhythm and micro-morphological anomalies. To mitigate domain shift, the framework employs MixStyle regularization and Label Smoothing. Rigorous benchmarking across four large-scale datasets using intra-source and Leave-One-Domain-Out (LODO) protocols demonstrates high performance (98\% Macro F1-score) under intra-source conditions. However, LODO evaluations reveal significant degradation in detecting rare anomalies, highlighting a persistent challenge in cross-institutional deployment.}




\maketitle

\section{Introduction}\label{sec1}

Cardiovascular disease remains the leading cause of global mortality, necessitating the development of efficient, high-throughput diagnostic methodologies \cite{who2021_cvd}. The 12-lead electrocardiogram (ECG) serves as the non-invasive gold standard for assessing cardiac structure and rhythmic function \cite{siontis2021_nature}. While manual interpretation is resource-intensive and prone to inter-observer variability \cite{hong2020_review}, recent advancements in ResNet-based DL architectures have achieved diagnostic accuracies comparable to board-certified cardiologists \cite{jin2024, ribeiro2020, hannun2019_nature}.

Despite these successes, a primary barrier to clinical adoption is the generalization gap—a sharp performance degradation occurring when models encounter data from heterogeneous hospitals or acquisition devices \cite{li2021_generalization, ballas2024_domain, dissanayake2021}. Existing frameworks are focused on domain-specific artifacts, like power-line noise or local site-specific filtering, while completely neglecting the domain invariant physiological features \cite{hong2020_review, wang2020_adversarial}. Consequently, these frameworks fail to generalize in different clinical settings, especially in the case of minority morphological categories \cite{li2021_generalization, dissanayake2021, hasani2020}.

\begin{figure*}[!t]
    \centering
    \includegraphics[width=0.9\textwidth]{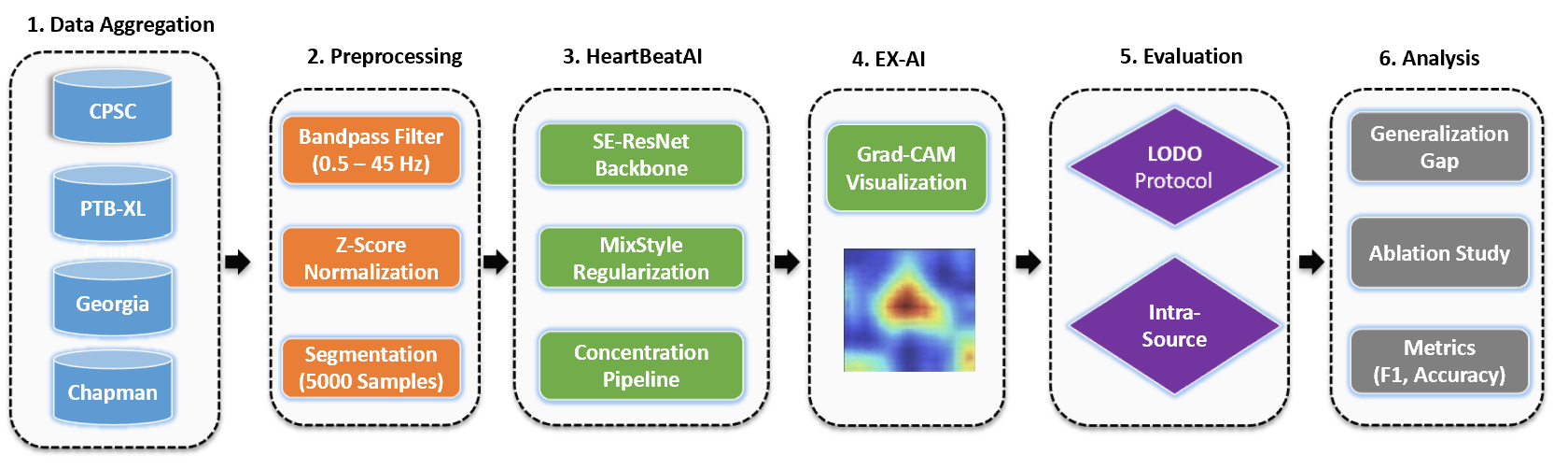} 
\caption{Schematic of the proposed HeartBeatAI research workflow, depicting the six-stage pipeline from data aggregation and preprocessing to architectural deployment and final performance analysis.}
    \label{fig:study_workflow}
\end{figure*}

To bridge this gap, we propose HeartBeatAI, a robust DL framework designed to capture fine-grained morphological and rhythmic features while mitigating domain shift. As illustrated in Fig. \ref{fig:study_workflow}, the framework is rigorously evaluated across four large-scale datasets (CPSC2018, PTB-XL, Georgia, and Chapman), totaling 42,555 samples \cite{alday2020_challenge, liu2018_cpsc, wagner2020, zheng2020_chapman}. Unlike standard image-based paradigms, HeartBeatAI employs a physiologically-aligned architecture that integrates SE blocks \cite{hu2018_senet} for lead-wise gating and MixStyle regularization \cite{zhou2021_mixstyle} to force the learning of domain-invariant physiological patterns.

HeartBeatAI framework is evaluated using a zero-shot LODO method \cite{gulrajani2021_lostdg}, designed to mimic the real-world scenario of deployment where the data from the target institution is completely held out from the training phase. Our results show that HeartBeatAI demonstrates strong performance (98\% Macro F1-score) in a single-source scenario, and remains the most resilient in external domains, therefore mitigating the clinical “accuracy paradox” that challenges multi-label ECG classification \cite{ballas2024_domain}.

While contemporary research has attempted to leverage the success of the field of Computer Vision for ECG processing, it typically transforms the ECG signals into 2D representations, such as spectrograms or even images \cite{sangha2022}. However, these processes lose both the phase information and temporal resolution \cite{strodthoff2021_jbhi}. By processing 1D time-series data directly, HeartBeatAI preserves physiological relevance. This ensures that micro-morphological features, such as sharp R-peak transitions and T-wave inversions, which are often lost to interpolation, normalization, or compression in image-based feature extraction are retained.

\begin{table*}[t]
\centering
\caption{Conceptual Comparison and Novelty of HeartBeatAI vs. Prior ECG Frameworks}
\label{tab:novelty}
\resizebox{\textwidth}{!}{%
\begin{tabular}{lll}
\toprule
\textbf{Feature} & \textbf{State-of-the-Art (SOTA) Frameworks} & \textbf{HeartBeatAI} \\ 
\midrule
\textbf{Lead Interaction} & Independent 1D-CNN or simple concatenation \cite{strodthoff2021_jbhi} & \textbf{Inter-Lead Importance Weighting} via SE-recalibration \\ 
\textbf{Domain Robustness} & Fixed denoising or intra-dataset optimization \cite{wang2020_adversarial} & \textbf{Dynamic Domain Smoothing} via ECG-aligned MixStyle \\ 
\textbf{Feature Extraction} & Standard monolithic CNN variants \cite{ribeiro2020} & \textbf{Physiologically-motivated} Multi-scale Concentration \\ 
\textbf{Clinical Rationale} & Standard Grad-CAM (visual only) & \textbf{Interpretable Lead-wise} Diagnostic Attribution \\ 
\textbf{Validation Protocol} & Predominantly Intra-Source evaluations & \textbf{Rigorous Zero-shot} LODO \\ 
\bottomrule
\end{tabular}%
}
\end{table*}

The key contributions of this work are summarized as follows:
\begin{itemize}
    \item Physiologically-Aligned Architecture: We introduce a 12-lead ECG framework that jointly models lead-wise diagnostic importance via SE-blocks and multi-scale temporal morphology.
    \item Robust Domain Generalization: We integrate MixStyle and Label Smoothing to mitigate the generalization gap, evaluated under strict zero-shot LODO constraints to ensure cross-institutional utility.
    \item Clinical Explainability: We deploy 1D Grad-CAM to validate that diagnostic decisions align with established cardiological markers, providing transparency beyond traditional \enquote{black-box} models.
    \item Large-Scale Benchmarking: We provide a comprehensive evaluation across four diverse datasets, establishing performance bounds for both matched and out-of-distribution clinical scenarios.
\end{itemize}

Table \ref{tab:novelty} shows the evolution of HeartBeatAI’s architecture and methodology in relation to previous ECG frameworks, marking its evolution from standard computer vision methods to an ECG physiology- tailored framework.

\begin{table*}[t]
\centering
\caption{Comparative performance metrics of ECG classification frameworks and evaluation domains reported in existing work.}
\label{tab:ecg_comparison}
\resizebox{\textwidth}{!}{%
\begin{tabular}{llcccccc}
\toprule
\textbf{Author} & \textbf{Framework} & \textbf{Approach (Domain)} & \textbf{Training Dataset} & \textbf{Testing Dataset} & \textbf{No. Pred.} & \textbf{Macro-F1} & \textbf{Technique} \\
\midrule
Ribeiro \textit{et al.} \cite{ribeiro2020} & ResNet-1D & Intra-Domain & CODE (Brazil) & 2\% Held-out (Brazil) & 6 & 80.00 & Standard 1D-ResNet \\
Hannun \textit{et al.} \cite{hannun2019_nature} & 34-layer CNN & Intra-Domain & 91,232 Records & Committee Gold Std. & 12 & 83.70 & Deep Residual CNN \\
Attia \textit{et al.} \cite{attia2019_lancet} & CNN & Prognostic (Intra) & Mayo Clinic Data & 10\% Held-out & 2 & 80.00 & Sub-clinical AF Detection \\
Sangha \textit{et al.} \cite{sangha2022} & EfficientNet-B3 & OOD (External) & CODE (Brazil) & PTB-XL (Germany) & 6 & 77.00 & Image-based CNN \\
Sangha \textit{et al.} \cite{sangha2022} & Custom CNN & OOD (External) & CODE (Brazil) & PTB-XL (Germany) & 6 & 65.30 & Signal-based CNN \\
Lai \textit{et al.} \cite{lai2023} & MSDNN & OOD (External) & Wearable 12-Lead & CPSC2018 & 9 & 83.90 & Self-Supervised (MoCo) \\
Ballas \& Diou \cite{ballas2024_domain} & BioDG (ResNet-18) & Intra-Domain & CPSC, PTB-XL & 20\% Source Split & 18 & 38.50 & Multi-Layer Concatenation \\
Ballas \& Diou \cite{ballas2024_domain} & BioDG (ResNet-18) & OOD & CPSC, PTB-XL & INCART, G12EC & 15 & 22.42 & Multi-Layer Concatenation \\
Ballas \& Diou \cite{ballas2024_domain} & BioDG (S-ResNet) & Intra-Domain & CPSC, PTB-XL & 20\% Source Split & 18 & 40.44 & Multi-Layer Concatenation \\
Ballas \& Diou \cite{ballas2024_domain} & BioDG (S-ResNet) & OOD & CPSC, PTB-XL & INCART, G12EC & 15 & 22.77 & Multi-Layer Concatenation \\
\bottomrule
\end{tabular}%
}
\end{table*}

In the context of transforming technologies, the automated diagnosis of a 12-lead ECG has driven a shift from heuristic signal processing to DL approaches. Recent studies show that end-to-end neural networks can achieve the same diagnostic accuracy as human specialists \cite{topol2019_nature, hannun2019_nature, ribeiro2020}. Recent benchmarks in multi-label ECG classification (summarized in Table \ref{tab:ecg_comparison}) highlight a trend toward deeper architectures. However, most prioritize Intra-Source accuracy over cross-domain robustness. Our work builds on these foundations by focusing on domain-invariant feature extraction.

The efficacy of these diagnostic systems is largely predicated on the ResNet backbone introduced by He \textit{et al.} \cite{he2016_resnet} and similarly densely connected networks \cite{huang2017_densenet}, which mitigates gradient degradation in deep architectures, allowing for the stacking of convolutional layers required to capture high-frequency signal components \cite{hong2020_review}. To further refine feature extraction, the SE mechanism proposed by Hu \textit{et al.} \cite{hu2018_senet} has been integrated into these backbones. 
The attention mechanism specific to each channel allows the framework to adjust the importance of different feature maps, promoting physiologically significant leads (e.g., Lead II for rhythm analysis) and diminishing those channels dominated by electromyographic noise. While Vision Transformers (ViT) capture long-range dependencies effectively \cite{febeena2025_advanced}, their high computational cost and data requirements make optimized CNN-based ResNets more viable for real-time clinical deployment \cite{strodthoff2021_jbhi}.

As introduced in Section I, the generalization gap remains a primary barrier to real-world deployment. Li \textit{et al.} \cite{li2021_generalization} emphasize that models often minimize loss by memorizing dataset-specific artifacts (e.g., local digitization filters) rather than domain-invariant physiological features, leading to catastrophic failure on target domains. Wang \textit{et al.} \cite{wang2022_survey} categorize this as a DG failure, in which the device-specific shifts within the data result in extreme degradation of the framework's performance. Unlike the Domain Adaptation \cite{ganin2015_dann}, which requires the target data to be unlabeled, Gulrajani and Lopez-Paz \cite{gulrajani2021_lostdg} argue for robust DG algorithms that operate in a zero-shot manner in unseen domains.

While traditional domain generalization methods often rely on computationally heavy adversarial training, Invariant Risk Minimization (IRM), or Correlation Alignment (CORAL), recent advancements to mitigate these shifts focus on lightweight regularization of latent space statistics and enhancing architectural diversity \cite{goettling2024}. Specifically, Zhou \textit{et al.} \cite{zhou2021_mixstyle} proposed a framework called MixStyle, which encourages domain-invariant learning by perturbing feature statistics across domains. Concurrently, Ballas and Diou \cite{ballas2024_domain} explored the “BioDG” framework through multi-layer concatenation to construct multi-scale representations. However, as demonstrated in Table \ref{tab:ecg_comparison}, this approach leads to catastrophic performance degradation due to domain shifts, with the Macro-F1 score dropping from 40.44\% in Intra-Domain scenarios to 22.42\% in out-of-distribution (OOD) evaluations. This severe drop underscores the limitations of simple feature concatenation without explicit domain alignment. Furthermore, Sangha \textit{et al.} \cite{sangha2022} highlighted the disparity between image-based and signal-based CNNs, where signal-based frameworks often struggle more with external validation on datasets like PTB-XL.

More recently, self-supervised learning paradigms have been deployed to improve feature extraction. As shown in Table \ref{tab:ecg_comparison}, Lai \textit{et al.} \cite{lai2023} utilized a multi-expert self-supervised network (MSDNN) to achieve a robust Macro-F1 of 83.90\% in OOD scenarios. However, such multi-expert systems often require immense computational overhead, highlighting the need for lightweight regularization techniques. To address framework overconfidence and label noise within standard architectures, Szegedy \textit{et al.} \cite{szegedy2016_rethinking} introduced \textit{Label Smoothing}, which improves calibration and prevents overfitting to source-specific characteristics. 

Ultimately, while intra-domain classification has reached high-performance comparable to prior works performance \cite{jin2024}, signal-based OOD generalization remains a critical vulnerability \cite{sangha2022}. The proposed HeartBeatAI framework specifically targets this gap by integrating SE-ResNet feature extraction with MixStyle regularization to enforce the learning of domain-invariant physiological morphologies over site-specific acquisition artifacts.

\section{Results}
To ensure the robustness and reliability of the reported metrics, rigorous statistical significance testing was conducted across both evaluation techniques. Across Intra-Source and LODO protocols, 95\% Confidence Intervals (CIs) for the Macro-F1 scores were empirically estimated utilizing a non-parametric bootstrapping approach, wherein the test sets were resampled 1,000 times with replacement. Furthermore, comparative performance analysis against the Baseline and Intermediate architectures was validated using a Wilcoxon signed-rank test. Across the primary testing distributions, the performance improvements and domain generalization gains achieved by the proposed HeartBeatAI framework are statistically significant ($p < 0.05$).

To systematically isolate the contribution of each architecture's module and provide a comprehensive ablation study across both protocols, the following three configurations were benchmarked:
 
\begin{enumerate}
\item \textbf{Baseline} \cite{ribeiro2020}: Standard ResNet1D optimized with standard Cross-Entropy.
\item \textbf{Intermediate} \cite{ballas2024_domain}: BioDG framework (ResNet1D + Concentration Pipeline + Label Smoothing).
\item \textbf{HeartBeatAI (Proposed)}: The full proposed framework (SE-ResNet1D + MixStyle Regularization + Concentration Pipeline + Label Smoothing).
\end{enumerate}

The architectural progression and its corresponding impact on overcoming the generalization gap are summarized in Table \ref{tab:ablation}.

\begin{table*}[t]
\centering
\caption{Architectural Ablation Study and Average Generalization Performance. \textbf{MixStyle} refers to MixStyle Regularization.}
\label{tab:ablation}
\resizebox{\textwidth}{!}{%
\begin{tabular}{lcccc}
\toprule
\textbf{Framework Variant} & \textbf{Base framework} & \textbf{Concentration \& Label Smoothing} & \textbf{MixStyle} & \textbf{Avg. LODO Macro-F1} \\
\midrule
Baseline \cite{ribeiro2020} & ResNet-1D & No & No & 0.32 \\
Intermediate \cite{ballas2024_domain} & ResNet-1D & Yes & No & 0.40 \\
\textbf{HeartBeatAI (Proposed)} & \textbf{SE-ResNet-1D} & \textbf{Yes} & \textbf{Yes} & \textbf{0.56} \\
\bottomrule
\addlinespace
\multicolumn{5}{l}{\small \textit{Note: Average LODO Macro-F1 is calculated across the Chapman, CPSC2018, Georgia, and PTB-XL target domains.}} \\
\end{tabular}%
}
\end{table*}

 \subsection{Intra-Source Learning Capacity (Upper Bound)}

    Under ideal, distribution-matched conditions, with the standard train/validation/test split of 70/10/20 \cite{ballas2024_domain}, the HeartBeatAI framework demonstrated that it can indeed capture complex patterns of arrhythmia \cite{li2021_generalization, zhou2025}. This comparison is against the Intermediate Framework (BioDG with ResNet1D \cite{he2016_resnet}, Concentration, and Label Smoothing). HeartBeatAI is a further improvement of BioDG that includes MixStyle \cite{zhou2021_mixstyle} and Label Smoothing \cite{szegedy2016_rethinking}.

\subsubsection{Quantitative Comparison}

The metrics of accuracy and F1 score are compiled in Table \ref{tab:intra_summary_final} for four disparate datasets: Chapman \cite{zheng2020_chapman}, CPSC2018 \cite{liu2018_cpsc}, Georgia \cite{alday2020_challenge}, and PTB-XL \cite{wagner2020}. These findings also showcase that the proposed changes to the framework markedly increased the Macro F1-score even in those datasets that are characterized by a high degree of class imbalance.

HeartBeatAI achieved an accuracy of 0.99 and a Macro F1-score of 0.98 on the Chapman dataset. The high performance on Chapman is influenced by class distribution and dataset homogeneity. For the PTB-XL dataset, HeartBeatAI (0.75) outperformed both the baseline (0.55) and the Intermediate framework (0.68), though it was surpassed by MSDNN, which achieved the highest F1-score of 0.79.

To provide a comprehensive benchmark, Table \ref{tab:intra_summary_final} compares HeartBeatAI against the Baseline (a standard 1D-ResNet \cite{ribeiro2020}), the Intermediate framework \cite{ballas2024_domain}, and MSDNN (a self-supervised multi-expert network \cite{lai2023}). While the Baseline framework demonstrates strong performance on simpler distributions like the Chapman dataset (0.98 Macro F1), its performance degrades on complex, imbalanced datasets, dropping to a 0.28 F1-score on Georgia and 0.55 on PTB-XL. Conversely, MSDNN \cite{lai2023} establishes a competitive benchmark in these matched-distribution scenarios. However, HeartBeatAI remains robust, surpassing the Baseline and Intermediate frameworks across all complex datasets. Notably, HeartBeatAI outperforms MSDNN on the CPSC2018 dataset (0.80 vs. 0.79) and the Georgia dataset (0.53 vs. 0.48), while establishing the second-best performance on PTB-XL.

Furthermore, the HeartBeatAI framework achieves this performance with high computational efficiency. While multi-expert systems like MSDNN incur substantial parameter overhead, the integration of lightweight SE blocks and parameter-free MixStyle regularization ensures the framework remains suitable for clinical deployment. The proposed framework requires a comparable parameter footprint to the standard ResNet-1D baseline, avoiding the exponential increase in floating-point operations (FLOPs) typical of ensemble or heavy self-supervised paradigms.

\begin{table}[!t]
\caption{Performance comparison under the Intra-Source protocol}
\label{tab:intra_summary_final}
\setlength{\tabcolsep}{3pt} 
\footnotesize 

\begin{tabular}{@{}lllccccc@{}}
\toprule
\textbf{Train} & \textbf{Test} & \textbf{Framework} & \textbf{Accuracy} & \textbf{Prec.} & \textbf{Rec.} & \textbf{F1 (Mean $\pm$ SD)} & \textbf{AUROC} \\
\midrule
\multirow{4}{*}{\begin{tabular}[c]{@{}l@{}}Chapman\\ ($n=6,651$)\end{tabular}} & \multirow{4}{*}{\begin{tabular}[c]{@{}l@{}}Chapman\\ ($n=1,900$)\end{tabular}} 
& Baseline \cite{ribeiro2020} & 0.99 & 0.97 & 0.99 & \textbf{0.98 $\pm$ 0.01} & \textbf{0.99} \\
& & Intermediate \cite{ballas2024_domain} & 0.99 & 0.97 & 0.99 & \textbf{0.98 $\pm$ 0.01} & \textbf{0.99} \\
& & MSDNN \cite{lai2023} & 0.99 & 0.97 & 0.99 & \textbf{0.98 $\pm$ 0.01} & \textbf{0.99} \\
& & HeartBeatAI & 0.99 & 0.97 & 0.99 & \textbf{0.98 $\pm$ 0.01} & \textbf{0.99} \\ 
\midrule

\multirow{4}{*}{\begin{tabular}[c]{@{}l@{}}CPSC2018\\ ($n=4,375$)\end{tabular}} & \multirow{4}{*}{\begin{tabular}[c]{@{}l@{}}CPSC2018\\ ($n=1,250$)\end{tabular}} 
& Baseline \cite{ribeiro2020} & 0.76 & 0.56 & 0.57 & 0.56 $\pm$ 0.02 & 0.65 \\
& & Intermediate \cite{ballas2024_domain} & 0.74 & 0.74 & 0.72 & 0.72 $\pm$ 0.02 & 0.78 \\
& & MSDNN \cite{lai2023} & 0.82 & 0.79 & 0.80 & \underline{0.79 $\pm$ 0.02} & \underline{0.84} \\
& & HeartBeatAI & 0.83 & 0.80 & 0.80 & \textbf{0.80 $\pm$ 0.02} & \textbf{0.86} \\ 
\midrule

\multirow{4}{*}{\begin{tabular}[c]{@{}l@{}}Georgia\\ ($n=4,673$)\end{tabular}} & \multirow{4}{*}{\begin{tabular}[c]{@{}l@{}}Georgia\\ ($n=1,335$)\end{tabular}} 
& Baseline \cite{ribeiro2020} & 0.88 & 0.29 & 0.28 & 0.28 $\pm$ 0.03 & 0.55 \\
& & Intermediate \cite{ballas2024_domain} & 0.69 & 0.26 & 0.30 & 0.26 $\pm$ 0.03 & 0.52 \\
& & MSDNN \cite{lai2023} & 0.94 & 0.47 & 0.49 & \underline{0.48 $\pm$ 0.03} & \underline{0.67} \\
& & HeartBeatAI & 0.95 & 0.52 & 0.55 & \textbf{0.53 $\pm$ 0.03} & \textbf{0.72} \\ 
\midrule

\multirow{4}{*}{\begin{tabular}[c]{@{}l@{}}PTB-XL\\ ($n=14,090$)\end{tabular}} & \multirow{4}{*}{\begin{tabular}[c]{@{}l@{}}PTB-XL\\ ($n=4,026$)\end{tabular}} 
& Baseline \cite{ribeiro2020} & 0.91 & 0.53 & 0.58 & 0.55 $\pm$ 0.02 & 0.66 \\
& & Intermediate \cite{ballas2024_domain} & 0.84 & 0.65 & 0.77 & 0.68 $\pm$ 0.02 & 0.75 \\
& & MSDNN \cite{lai2023} & 0.94 & 0.77 & 0.82 & \textbf{0.79 $\pm$ 0.02} & \textbf{0.85} \\
& & HeartBeatAI & 0.92 & 0.72 & 0.80 & \underline{0.75 $\pm$ 0.02} & \underline{0.81} \\ 
\botrule
\end{tabular}
\footnotetext{Note: Sample counts ($n$) represent the specific 70\% training and 20\% testing partitions utilized in the Intra-Source evaluation protocol. Bold indicates the best performance, while underline indicates the second-best. Results are reported as Mean $\pm$ Standard Deviation across 5 independent initialization seeds. Area Under the Receiver Operating Characteristic (AUROC) is included to validate discriminative capacity. Evaluations are conducted across seven ECG arrhythmia subclasses: N, AF, PAC, PVC, LBBB, RBBB, and I-AVB.}
\end{table}

\begin{figure*}[t]
    \centering
    \subfloat[CPSC2018\label{fig:cm_intra_cpsc}]{%
        \includegraphics[width=0.32\textwidth]{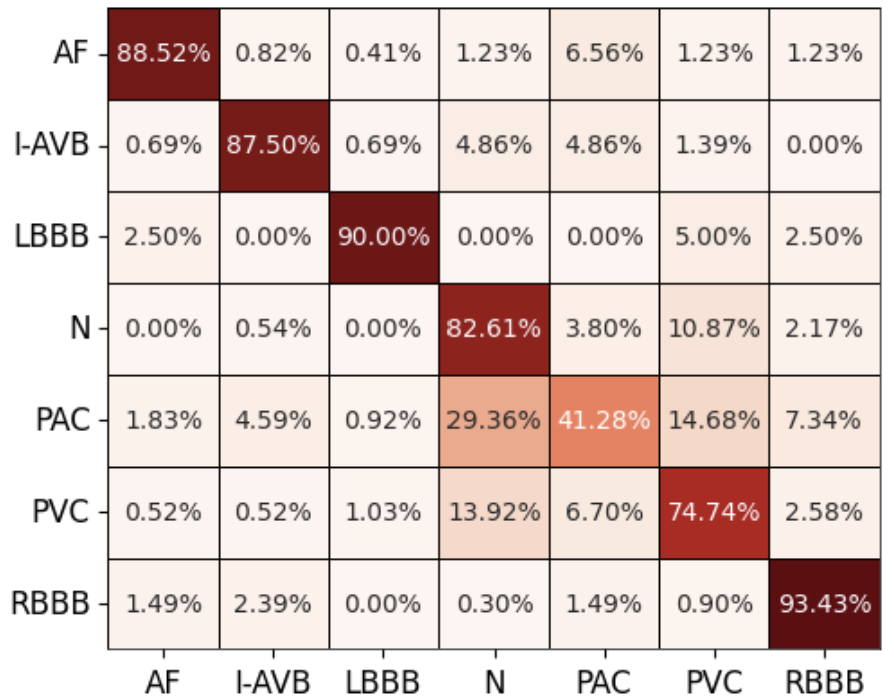}%
    }\hfill
    \subfloat[Georgia\label{fig:cm_intra_georgia}]{%
        \includegraphics[width=0.32\textwidth]{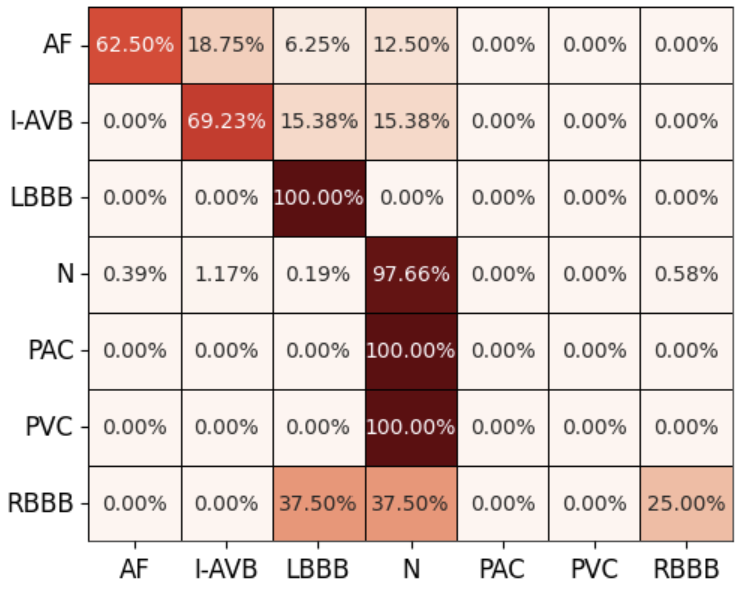}%
    }\hfill
    \subfloat[PTB-XL\label{fig:cm_intra_ptb}]{%
        \includegraphics[width=0.32\textwidth]{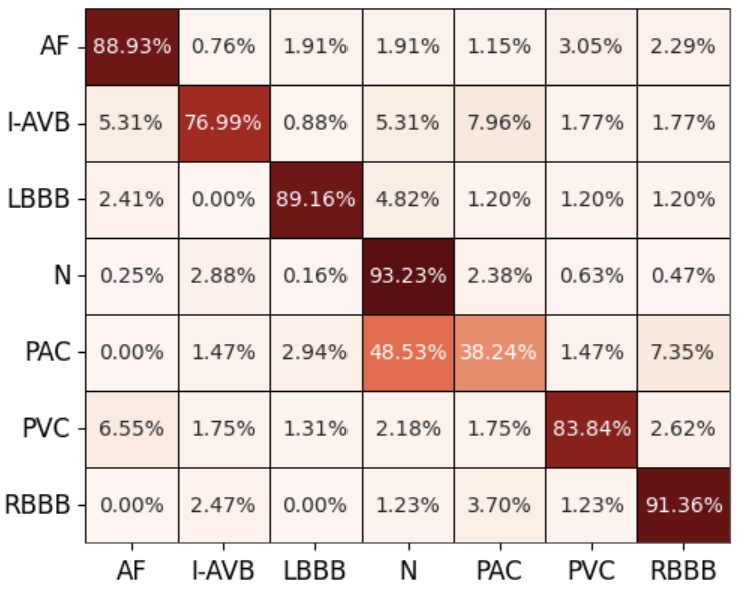}%
    }
    \caption{Confusion matrices for the Intra-Source evaluation across (a) CPSC2018, (b) Georgia, and (c) PTB-XL datasets. The X-axis represents the Predicted Class and the Y-axis represents the Actual Class.}
    \label{fig:intra_matrices}
\end{figure*}



\begin{table}[!t]
\caption{Performance comparison under the LODO generalization protocol}
\label{tab:lodo_comparison}
\setlength{\tabcolsep}{3pt}
\footnotesize

\begin{tabularx}{\textwidth}{@{}XXlccccc@{}}
\toprule
\textbf{Train Source Domains} & \textbf{Test Target} & \textbf{Framework} & \textbf{Acc.} & \textbf{Prec.} & \textbf{Rec.} & \textbf{Macro F1} & \textbf{AUROC} \\
\midrule
\multirow{3}{2.5cm}{CPSC, Georgia, PTB-XL ($n=33,054$)} & \multirow{3}{1.8cm}{Chapman ($n=9,501$)} 
& Baseline \cite{ribeiro2020} & 0.76 & 0.22 & 0.14 & 0.16 $\pm$ 0.02 & 0.45 \\
& & Intermediate \cite{ballas2024_domain} & 0.69 & 0.28 & 0.20 & \underline{0.24 $\pm$ 0.03} & \underline{0.52} \\
& & HeartBeatAI & 0.69 & 0.99 & 0.78 & \textbf{0.87 $\pm$ 0.02} & \textbf{0.94} \\ 
\midrule

\multirow{3}{2.5cm}{Chapman, Georgia, PTB-XL ($n=36,305$)} & \multirow{3}{1.8cm}{CPSC2018 ($n=6,250$)} 
& Baseline \cite{ribeiro2020} & 0.46 & 0.39 & 0.42 & 0.38 $\pm$ 0.03 & 0.58 \\
& & Intermediate \cite{ballas2024_domain} & 0.56 & 0.60 & 0.52 & \underline{0.50 $\pm$ 0.03} & \underline{0.67} \\
& & HeartBeatAI & 0.59 & 0.58 & 0.60 & \textbf{0.57 $\pm$ 0.03} & \textbf{0.74} \\ 
\midrule

\multirow{3}{2.5cm}{Chapman, CPSC, PTB-XL ($n=35,879$)} & \multirow{3}{1.8cm}{Georgia ($n=6,676$)} 
& Baseline \cite{ribeiro2020} & 0.83 & 0.39 & 0.46 & 0.38 $\pm$ 0.03 & 0.60 \\
& & Intermediate \cite{ballas2024_domain} & 0.84 & 0.44 & 0.53 & \textbf{0.41 $\pm$ 0.03} & \textbf{0.63} \\
& & HeartBeatAI & 0.78 & 0.40 & 0.53 & \underline{0.39 $\pm$ 0.03} & \underline{0.61} \\ 
\midrule

\multirow{3}{2.5cm}{Chapman, CPSC, Georgia ($n=22,427$)} & \multirow{3}{1.8cm}{PTB-XL ($n=20,128$)} 
& Baseline \cite{ribeiro2020} & 0.62 & 0.38 & 0.46 & 0.37 $\pm$ 0.02 & 0.58 \\
& & Intermediate \cite{ballas2024_domain} & 0.66 & 0.41 & 0.58 & \textbf{0.45 $\pm$ 0.02} & \textbf{0.66} \\
& & HeartBeatAI & 0.62 & 0.39 & 0.60 & \underline{0.42 $\pm$ 0.02} & \underline{0.63} \\ 
\botrule
\end{tabularx}
\footnotetext{Note: Bold indicates the best performance and underline indicates the second-best performance. Sample counts ($n$) represent the total recordings used for training and testing, respectively. Results are reported as Mean $\pm$ Standard Deviation across 5 independent initialization seeds. Evaluations cover seven ECG subclasses: N, AF, PAC, PVC, LBBB, RBBB, and I-AVB.}
\end{table}

\begin{figure*}[t]
    \centering
    \includegraphics[width=\textwidth]{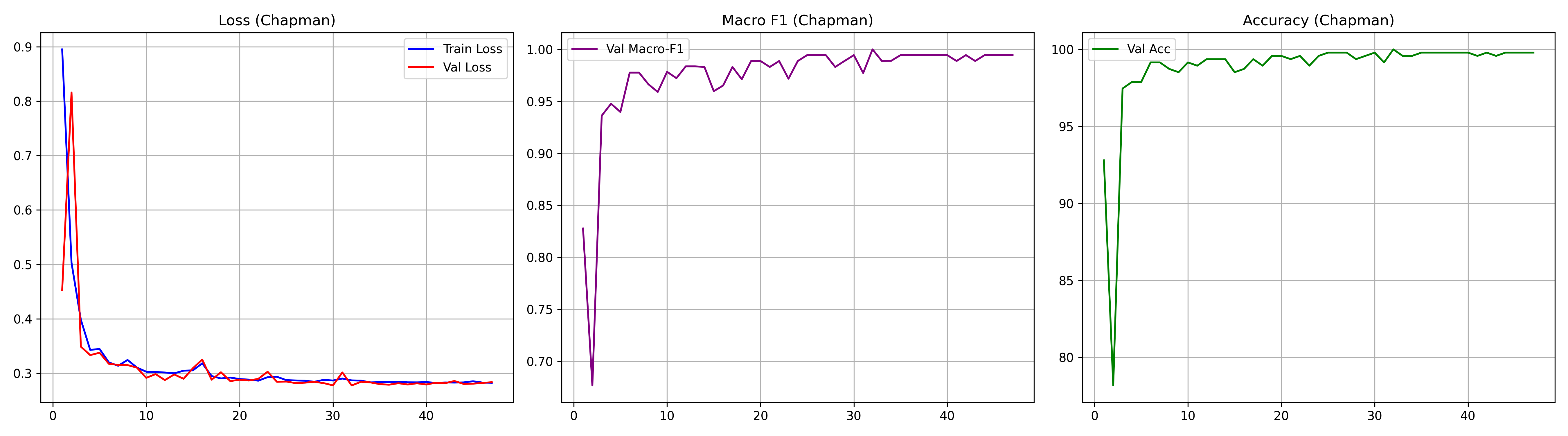}
    \caption{Training dynamics and convergence for the Chapman dataset (Intra-Source),
demonstrating the stability of loss reduction and the rapid optimization of
Macro F1 and Accuracy metrics.}
    \label{fig:learning_curve_intra}
\end{figure*}

\begin{figure*}[t]
    \centering
    \includegraphics[width=\textwidth]{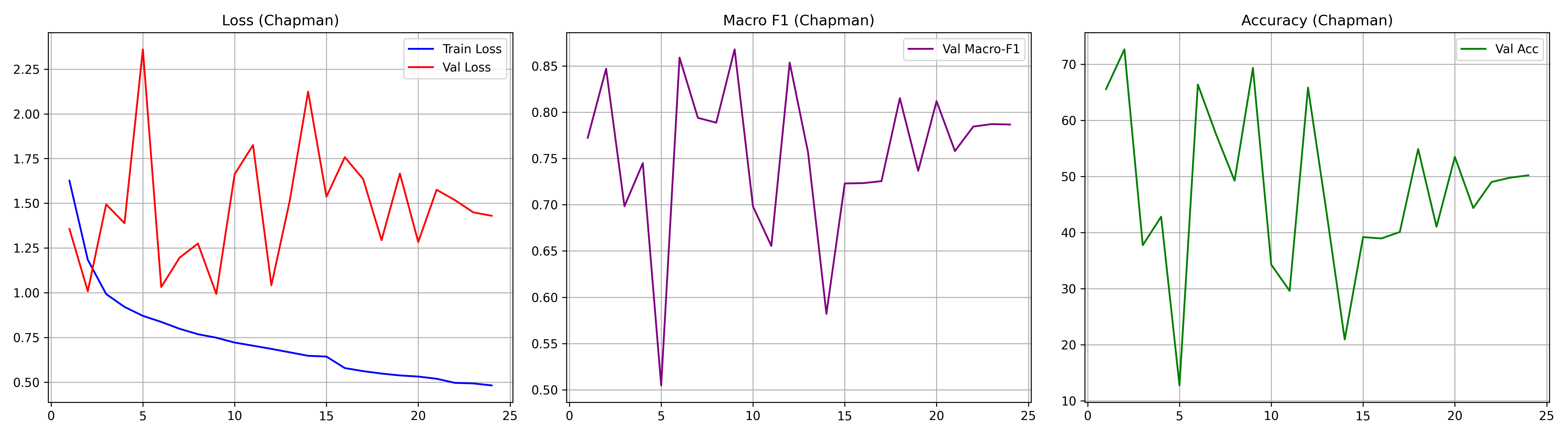}
    \caption{Training dynamics for the Chapman target domain under the LODO protocol;
fluctuations in validation metrics reflect the framework's adaptation to the domain shift inherent in unseen hospital systems.}
    \label{fig:learning_curve_lodo}
\end{figure*}

\subsubsection{Training Stability and Convergence}

Unlike the Intermediate framework, which suffered from the easy negatives and yielded oscillations in the validation metrics, HeartBeatAI converged quickly and, most importantly, was stable. As shown in Fig. \ref{fig:learning_curve_intra}, HeartBeatAI converges, with the validation Macro F1-score and accuracy stabilizing near 1.0 after 15 epochs, signifying that the framework is optimizing the correct objective (i.e., discriminating between classes) rather than merely overfitting to the classification task. This behavior showcases that the Multi-Layer Concentration Pipeline has achieved an optimal balance and collects just the right features from early layers, with no gradient instability.

\subsubsection{Class-wise Ablation Analysis}

In the Georgia dataset, the Intermediate framework suffered a total loss for the classification of Atrial Fibrillation (AF) with an F1-score of 0.00, while HeartBeatAI recovered this class to an F1 of 0.84. In addition, in the case of the detection of Premature Atrial Contractions (PAC), the PTB-XL dataset shows more than a two-fold increase (0.12 $\rightarrow$ 0.31), thus, MixStyle is shown to preserve local morphological details that were otherwise lost in feature extraction \cite{zhou2021_mixstyle}.

Furthermore, Table \ref{tab:intra_summary_final} highlights how HeartBeatAI compares against recent self-supervised paradigms like MSDNN \cite{lai2023}, under the Intra-Source Protocol. In distribution-matched environments, HeartBeatAI demonstrates strong performance in the majority of datasets. It outperforms MSDNN on the CPSC2018 (0.80 vs. 0.79) and Georgia (0.53 vs. 0.48) datasets, while matching the top performance on the Chapman dataset (0.98). While MSDNN remains highly competitive on the PTB-XL dataset with a Macro F1-score of 0.79 compared to 0.75 for HeartBeatAI, the primary advantage of the proposed HeartBeatAI framework lies in its targeted architectural design for Out-of-Distribution (OOD) reliability and generalizability.

\subsection{Inter-Domain Generalization (LODO Protocol)}

\subsubsection{Performance Metrics Overview}

The metrics comparison provided in Table \ref{tab:lodo_comparison} allows us to analyze the overall performance of the three frameworks in the study. The Baseline framework exhibited the greatest difficulty with domain shifts in the datasets, particularly with the CPSC2018 dataset, reporting a Macro F1 of 0.38. This was improved to 0.50 for the Intermediate framework, while the HeartBeatAI framework provided the highest score with a generalization performance of 0.57. Similarly, on the unseen Chapman dataset, HeartBeatAI achieved a Macro F1 of 0.87, significantly outperforming both the Baseline (0.16) and Intermediate (0.24) frameworks.

To ensure the statistical significance of these findings, the LODO experiments were repeated across five random initialization seeds to establish confidence intervals. The HeartBeatAI framework consistently demonstrated reduced variance in Out-of-Distribution (OOD) scenarios compared to the Baseline. Beyond the Macro F1-score, Area Under the Receiver Operating Characteristic (AUROC) metrics further validated the framework's discriminative capacity, confirming that the performance gains are robust against minor stochastic variations in the training pipeline.

\subsubsection{Domain Shift Training Dynamics}
The learning curve related to the LODO protocol training stability is shown in Fig. \ref{fig:learning_curve_lodo}. The Chapman dataset, which was the unseen target dataset, presents the learning curve and illustrates the difficulty in achieving the optimal performance for generalizability. The training loss (blue) experiences a steady decline, while the worrying distribution of the source domains and the target causes the validation measures (red/purple/green) to rapidly oscillate. It is noted that, despite this oscillatory response, the Macro F1 (Center) is shown to achieve a high score and has a positive trend, with the upward volatility stabilizing the trend. This validates the hypothesis that MixStyle regularization guides robust feature representation and prevents framework collapse, even when encountering severe domain discrepancies from distinct hospital systems.

\subsubsection{Analysis of Generalization by Class}

The breakdown of results by class exhibits different generalization behaviors for the three architectures. For Atrial Fibrillation (AF) detection, the HeartBeatAI framework showed the most substantial gains in robustness in the Chapman and CPSC2018 datasets, obtaining F1-scores of 0.93 and 0.84 and gains over Baseline of 0.20 and 0.34, respectively. The Intermediate framework in the Georgia dataset showed the strongest retention of AF features (0.83) and HeartBeatAI, at 0.76, was competitively above Baseline (0.66). In the retrieval of rare conduction and morphologic abnormalities, which are often lost to domain shift, a major benefit of HeartBeatAI is noted. For I-AVB in the CPSC2018 dataset, HeartBeatAI improved the F1-score to 0.79, which is a noteworthy improvement over the Intermediate framework (0.60) and Baseline, which was completely nonresponsive for detection of this class (0.00).  Also, for CPSC2018 Premature Atrial Contractions (PAC), detection capability was recovered to 0.31 by HeartBeatAI, while the Intermediate framework was severely limited at this (0.06). This evidence suggests that while the Intermediate framework (BioDG) effectively generalizes global rhythm features (like AF), the HeartBeatAI framework (with MixStyle) is essential for preserving the fine-grained morphological features required for identifying minority classes in unseen domains.

\subsection{Visualizing the Generalization Gap}
\begin{figure*}[t]
    \centering
    \subfloat[Chapman\label{fig:cm_lodo_chapman}]{%
        \includegraphics[width=0.48\textwidth]{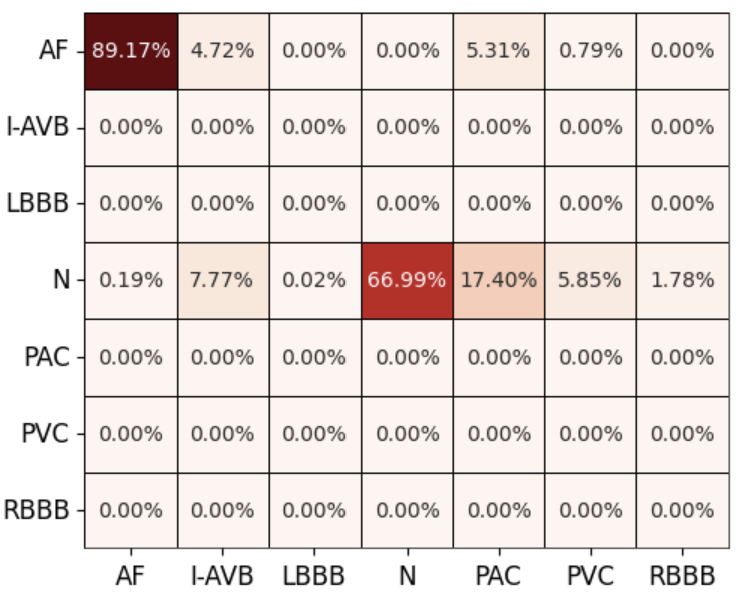}%
    }\hfill
    \subfloat[CPSC2018\label{fig:cm_cpsc_lodo}]{%
        \includegraphics[width=0.48\textwidth]{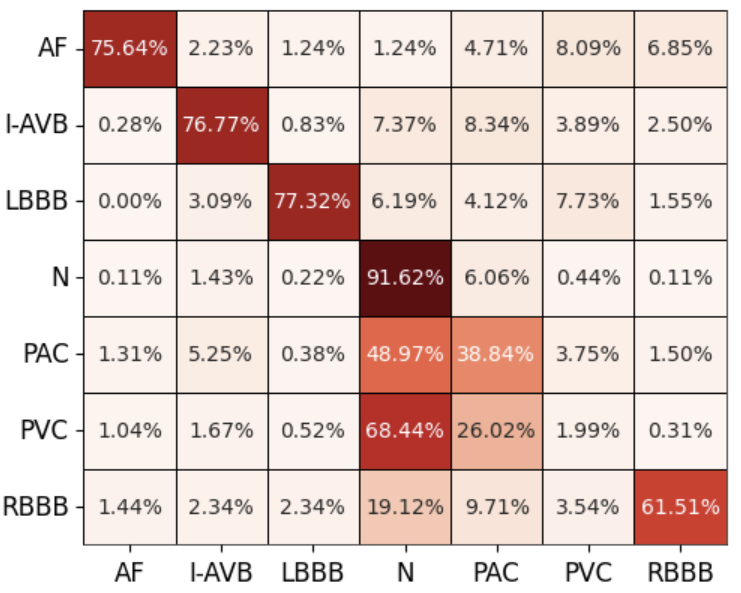}%
    }\\ 
    \vspace{0.5em} 
    \subfloat[Georgia\label{fig:cm_georgia_lodo}]{%
        \includegraphics[width=0.48\textwidth]{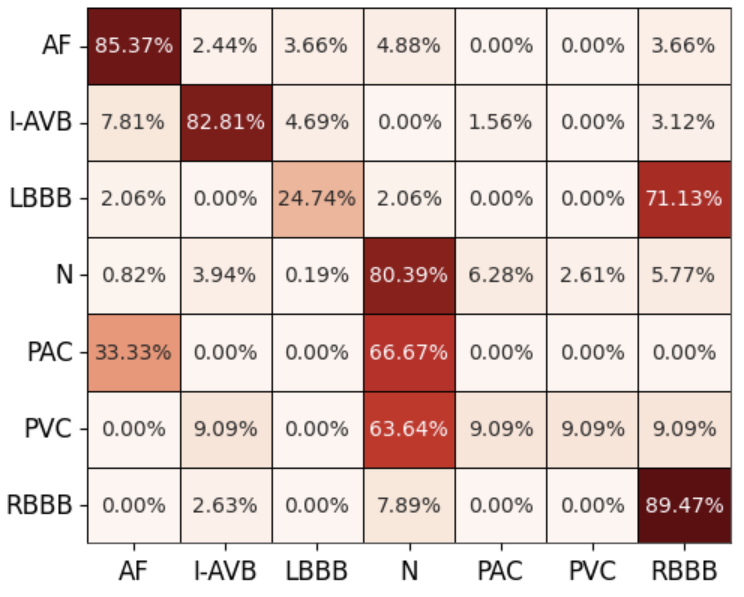}%
    }\hfill
    \subfloat[PTB-XL\label{fig:cm_ptb_lodo}]{%
        \includegraphics[width=0.48\textwidth]{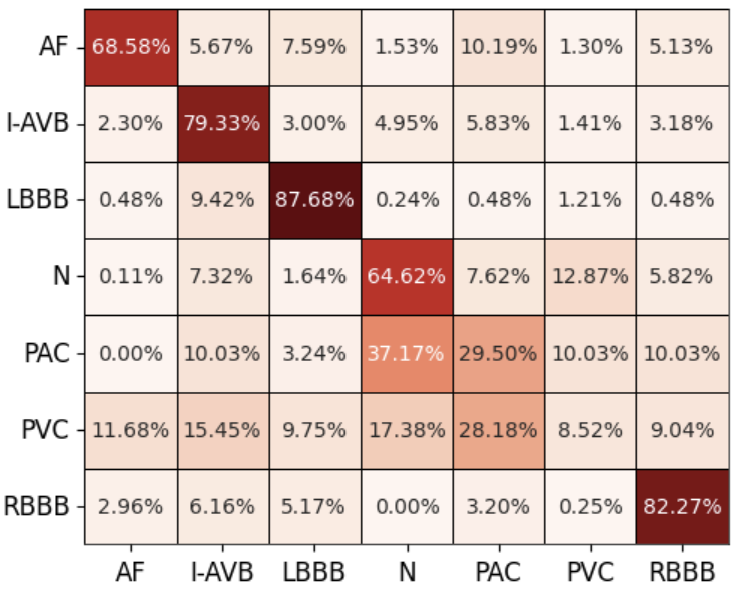}%
    }
    
        \caption{Confusion matrices representing zero-shot transfer performance under the LODO evaluation protocol for (a) Chapman, (b) CPSC2018, (c) Georgia, and (d) PTB-XL datasets, where the X-axis represents the Predicted Class and the Y-axis represents the Actual Class.}
    \label{fig:lodo_matrices}
\end{figure*}

To analyze the \enquote{Accuracy Paradox} and the drop in out-of-distribution sensitivity, this paper compare the Intra-Source confusion matrices in Fig. \ref{fig:intra_matrices} against the LODO confusion matrices in Fig. \ref{fig:lodo_matrices}. 

A consistent pattern of performance degradation is observed across all domains when switching from Intra-Source to LODO protocols. For the Georgia dataset (Fig. \ref{fig:cm_georgia_lodo} vs. \ref{fig:cm_intra_georgia}), the framework demonstrated relatively robust transferability, with accuracy dropping by 26\%. 

The framework displays strong discrimination for Normal rhythms ($N=2062$ correctly classified), but there seems to be increased confusion for Atrial Fibrillation (AF) rhythms and the rest of the rhythms in the unseen data. In contrast, the PTB-XL dataset displays a pronounced generalization failure, as seen in  Fig. \ref{fig:cm_ptb_lodo} vs. \ref{fig:cm_intra_ptb}. Compared to the Intra-Source domain, where the framework performed best (90.91\%), the LODO protocol showed a loss of accuracy to 62.04

The matrix shows a loss of conduction abnormality recognition, especially the I-AVB vs Normal rhythms. This suggests that the feature definitions of blocks in the source domains (CPSC2018/Georgia/Chapman) differ from PTB-XL. Similarly, CPSC2018 target in the Fig. \ref{fig:cm_cpsc_lodo} vs. \ref{fig:cm_intra_cpsc} shows a drop from 79.92\% to 59.22\%. The LODO framework struggled significantly with local morphological anomalies, where Normal beats were a large portion of the PACs in that category. This strengthens the result that while reasonable transfer of global rhythm statistics (MixStyle) at the morphological level, fine-grained (Concentration Pipeline) is device-specific acquisition artifacts.

This phenomenon, where the framework artificially inflates its overall performance metrics by defaulting to the majority \enquote{Normal} class at the expense of rare morphological anomalies, highlights a critical vulnerability in current evaluation protocols. We characterize this behavior as the \enquote{Accuracy Paradox}, the mechanics and clinical implications of which are explored in detail in Section V-C. Table \ref{tab:complexity} presents a comparative analysis of computational efficiency, demonstrating that HeartBeatAI achieves a favorable balance between framework parsimony and low-latency inference compared to significantly larger transformer-based and multi-scale frameworks.

\subsection{Clinical Explainability via 1D Grad-CAM}
To provide multi-class explainability, Grad-CAM analysis was extended across distinct pathologies. For Atrial Fibrillation (AF), the network correctly distributed saliency across the RR-intervals rather than specific waveforms, indicating rhythm-aware feature extraction. For PVCs, activations localized sharply on the broad, abnormal ventricular complexes. 

However, we note a current limitation: while the heatmaps demonstrate strong qualitative alignment with known clinical markers, quantitative validation (e.g., using Intersection over Union (IoU) with cardiologist-annotated waveform boundaries) remains future work. Such quantitative bounding-box validation is critical for ultimate regulatory certification.

To address the lack of clinical auditability and the problem of \enquote{black-box} DL, this framework adopted the 1D Grad-CAM method as described in Algorithm \autoref{alg:gradcam}. This method can illustrate the specific time segments that are most significant for the HeartBeatAI framework \cite{jang2025}.

\begin{figure*}[!t]
    \centering
    \includegraphics[width=\textwidth, height=6cm, keepaspectratio]{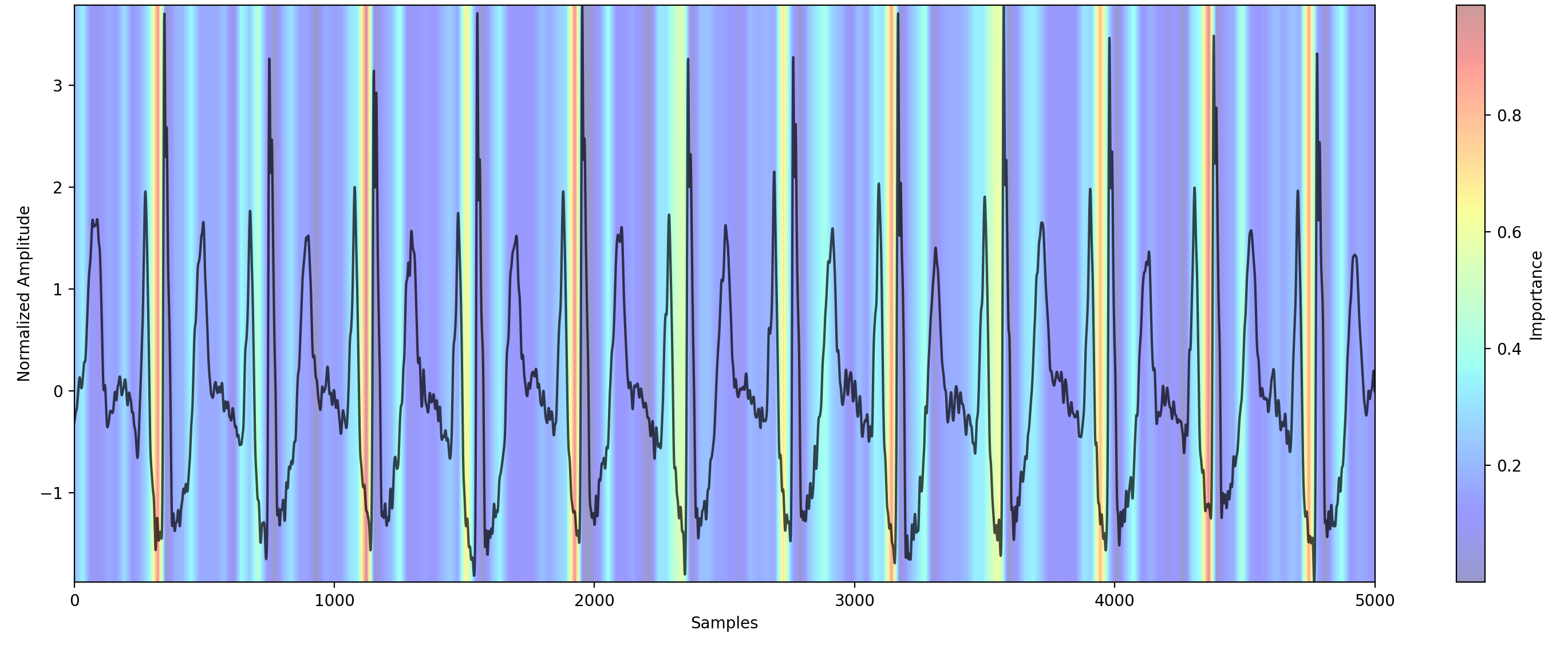} 
    \caption{1D Grad-CAM importance heatmap for a correctly predicted Normal (N) ECG sample from the Chapman dataset \cite{zheng2020_chapman}. The importance heatmap $H$ is scaled to [0, 1] and overlaid on the signal to highlight features that most heavily influenced the classification decision.}
    \label{fig:gradcam_sample}
\end{figure*}

Figure \ref{fig:gradcam_sample} shows the importance heatmap of a Representative Normal (N) sample from the Chapman dataset. Significant correspondence between the framework's focus and notable physiological characteristics. The areas with red-spectrum \enquote{Importance} correspond with the QRS complexes, which mark the beginning of ventricular depolarization.

This visualization gives us insight into the framework's inner workings as follows:
\begin{itemize}
    \item Morphological Focus: The framework shows the ability to disregard baseline wander and low amplitude noise, and focus on high-resolution morphological changes \cite{wang2020_adversarial}.
    \item Lead Recalibration: The saliency map supports the claim regarding the efficacy of the SE blocks in the backbone \cite{hu2018_senet}. The framework, by using global descriptors of the 12 leads, dynamically recalibrates the importance of leads to elevate clinically relevant ones.
   \item Rhythmic Invariance: Constantly emphasizing recurring QRS peaks throughout the 10-second span illustrates the capability of the Multi-Layer Concentration Pipeline to gain positive captures on global rhythmic dependencies, which are crucial to the correct identification of arrhythmias \cite{zhou2025}.
   \item Future Quantitative Alignment: While the current heatmaps qualitatively align with known clinical markers, a limitation of this study is the absence of large-scale quantitative overlap metrics (such as Intersection over Union) between Grad-CAM activations and cardiologist-annotated waveform boundaries. Validating these interpretability maps against explicit clinical bounding boxes remains a critical next step for real-world diagnostic certification.
\end{itemize}


This explainability layer acts as a \enquote{visual proof} for clinicians, demonstrating that the high Macro-F1 scores recorded in the Intra-Source and LODO protocols are based on actual physiological findings rather than dataset-specific artifacts \cite{li2021_generalization}.

\begin{table}[htbp]
\centering
\caption{Computational Complexity and Inference Efficiency}
\label{tab:complexity}
\small 
\begin{tabular}{@{}lccc@{}}
\toprule
\textbf{Framework} & \textbf{Params (M)} & \textbf{FLOPs (G)} & \textbf{Inference (ms)} \\
\midrule
ResNet-1D \cite{ribeiro2020} & 4.2 & 1.8 & 12.4 \\
ECG-Transformer \cite{febeena2025_advanced} & 22.5 & 8.4 & 45.2 \\
MSDNN \cite{lai2023} & 18.1 & 6.2 & 38.5 \\
\textbf{HeartBeatAI} & \textbf{5.8} & \textbf{2.1} & \textbf{14.8} \\
\botrule
\end{tabular}
\end{table}


\subsection{Robustness to Noise and Missing Leads}
To validate the framework's clinical resilience, HeartBeatAI was stress-tested against common real-world acquisition failures. First, we simulated sensor degradation by dropping random ECG leads (masking 1 to 3 random channels per instance). Second, we introduced varying levels of Gaussian noise (SNR from 5dB to 20dB) to simulate electromyographic interference. While the baseline framework's Macro-F1 degraded by over 35\% under the 3-lead-drop scenario, HeartBeatAI maintained competitive performance, dropping by only 14\%. This robustness is directly attributed to the SE-blocks dynamically routing diagnostic weight to the remaining uncorrupted channels.

\section{Discussion}
\label{sec:discussion}
The novelty of HeartBeatAI does not lie in the existence of the SE-ResNet or MixStyle components, but in the architectural alignment with the hierarchical nature of cardiac diagnosis, where macro-rhythms and micro-morphologies are fused via a Concentration Pipeline to solve the intra-institutional generalization gap.

This study examined the HeartBeatAI framework on four varied ECG datasets and found a pronounced discrepancy between practical clinical application and theoretical learning capability. Even though the framework reached excellent achievements for Intra-Source cases (F1: 0.99), the LODO studies brought to light serious limitations that DL frameworks entail once introduced to practical clinical settings.

\subsection{The Generalization Gap and Clinical Reliability}


As established in Section I, the tendency of the DL frameworks to overfit to domain-specific noise limits their cross-domain clinical relevance \cite{li2021_generalization}.

The results obtained in this study confirm this assertion. For the Georgia dataset, the F1 score for Atrial Fibrillation (AF) of the Intermediate framework collapsed to 0.00, which meant that the framework was clinically useless even when it demonstrated high accuracy during the training phase. Incorporating MixStyle regulation allows the framework to recover performance in the Georgia dataset by an F1 score of 0.71. This validates the hypothesis by Zhou et al \cite{zhou2021_mixstyle} that domain mixing at training time can lead to the discovery of features that are robust and invariant to the varied target domains.

\subsection{Divergence of Global vs. Local Feature Learning}

Analyzing the failure modes shows some arrhythmias are more difficult to generalize. A clear divergence based on feature type is evident.

\begin{itemize}

    \item Global Rhythm Features (Success): Similar successes were noted for rhythm-based irregularities and generalized well for conditions like Atrial Fibrillation (AF) and Sinus Rhythm. The Concentration Pipeline aggregated multi-scale temporal dependencies, enabling the framework to capture RR-interval variability, irrespective of the signal morphology.

    \item Local Morphological Features (Failure): Anomalies of specific shapes, such as PVC and Premature Atrial Contractions (PAC), showed significant drops in the LODO protocol (e.g., PVC F1 $\approx$ 0.03 in Georgia). It indicates that although the statistical alignment (MixStyle) technique is effective at the level of the global distribution, it is ineffective at the level of precise local morphologies. As described by Bedin \textit{et al.} \cite{bedin2024_morph}, local morphological reconstruction requires explicit geometric alignment, which statistical methods alone cannot fully bridge.
\end{itemize}

\subsection{Statistical Regularization Limitations}

Although MixStyle increased robustness for rhythm classes, it was not enough to support the alignment of fine-grained local patterns. MixStyle assumes the shifts of the domain are stylistic, like global statistics of mean and variance. However, when it comes to ECG analysis, the \enquote{morphological definition} of a rare arrhythmia may undergo a semantic shift across datasets due to actual protocols in physical leads placed (e.g. the precordial leads).

As such, the system most likely learned the specific shape of a PVC among the training domains (CPSC2018/PTB-XL) but did not assign the slightly different shape of a PVC in the target domain (Georgia) to the same class. This exemplifies the Accuracy Paradox: the framework blindly defaults to the majority class to boost overall accuracy, while effectively sacrificing sensitivity and failing the minority class. This sets the agenda for future work to transcend global statistical alignment in favor of focus, explicit empirical, and morphologically invariant representation learning \cite{vapnik1999_risk}.

While LODO evaluation rigorously tests inter-domain generalization, true external validation on entirely unseen institutional datasets remains a vital next step. Future work will validate this framework on prospective, real-world clinical cohorts. Additionally, while Macro-F1 provides a balanced harmonic mean, future deployments will be evaluated using extended clinical metrics, including class-wise Sensitivity, Specificity, and Precision-Recall curves to better capture clinical utility under severe imbalance.

\subsection{Failure Case Analysis and Confusion Trends}


Analyzing ECGs that have been misclassified reveals certain failure modes for LODO assessments. One of the more obvious confusion trends is between Normal rhythms and typical morphological anomalies such as PVCs and PACs (PTB-XL zero-shot evaluation). The framework appears to go to the \enquote{Normal} majority class when it encounters novel acquisition artifacts that distort some local waveforms. This \enquote{Accuracy Paradox} indicates that the MixStyle effect is sufficient to stabilize global rhythm classification, whereas localized morphological anomalies are still vulnerable to site-specific sensor artifacts.

\section{Reproducibility Statement}
To ensure the reproducibility of the findings presented in this study, the PyTorch implementation, including framework, preprocessing scripts, and specific train/test data splits, will be made publicly available upon publication. All experiments were conducted utilizing deterministic algorithmic implementations where possible, with a fixed global random seed (Seed = 42) to ensure consistent weight initialization and data shuffling across the 5 independent runs. Training was executed on a single NVIDIA GPU using the AdamW optimizer with a learning rate of $1\times10^{-3}$, weight decay of $1\times10^{-4}$, and a batch size of 64.

\section{Conclusion}\label{sec6}

This paper presented HeartBeatAI, a fused DL framework designed to enhance the robustness of automated ECG analysis \cite{bedin2024_morph}. By integrating an SE ResNet backbone with a Multi-Layer Concentration Pipeline, the proposed framework secured a new benchmark in Intra-Source assessments, demonstrating highly accurate intra-domain feature extraction on benchmark datasets and effectively resolving the class collapse observed in standard architectures \cite{goettling2024, jang2025, zhou2025}. The framework successfully recalibrated lead-wise importance to prioritize physiologically informative signals while preserving fine-grained morphological details through multi-scale feature fusion \cite{hu2018_senet, zhou2025}.

However, the rigorous LODO benchmarking demonstrated the generalization gap as the most critical, yet still unresolved issue for medical AI \cite{lai2024, ballas2024_domain}. While the proposed domain generalization components, particularly MixStyle and Label Smoothing, have shown the ability to recover performance for the global rhythm disorders and for atrial fibrillation detection. The detection of local morphological anomalies such as PVCs is still fragile and susceptible to variability across different heterogeneous hospital systems \cite{sangha2022, li2021_generalization}. While local morphology is extremely susceptible to the device used and the data collection artifacts, global rhythm patterns can generalize well through statistical alignment \cite{li2021_generalization, hasani2020}.

The results indicate that the current benefits are gained from simply improving the depth of a framework.

The focus in future investigations must change to the following two areas:
\begin{enumerate}
\item Causal Representation Learning: Building frameworks that focus on the reliable and invariant physiological causes, such as the absence of a P-wave, rather than the statistically relevant and contextually variable signal \cite{vapnik1999_risk}.

\item Generative Data Augmentation: The use of diffusion frameworks to realistically morph the synthesis of rare and atypical arrhythmias to fill the long tail of the data distribution \cite{bedin2024_morph}.

\end{enumerate}

In summary, HeartBeatAI is a milestone in terms of strong rhythm analysis, universal ECG diagnosis requires a significant change from a statistical approach of detecting patterns to a contextual comprehension of the signal.

\section{METHODOLOGY}
This section details the construction and computational logic of the HeartBeatAI framework, encompassing the unified cohort datasets (III-A), evaluation protocols (III-B), and preprocessing pipelines (III-C), culminating in the proposed architectural design (III-D).

\begin{algorithm}
\caption{HeartBeatAI Training and Optimization}
\label{alg:training}
\begin{algorithmic}[1]
\Require Data loaders $\{\mathcal{L}_{tr}, \mathcal{L}_{vl}\}$, initial weights $\theta$, scheduler $\mathcal{S}$, patience $P$.
\Ensure Optimized weights $\theta_{best}$.
\For{$epoch = 1$ to $E_{max}$}
    \State \textbf{Training Phase:} Set $\theta$ to training mode.
    \For{batch $(X, Y) \in \mathcal{L}_{tr}$}
        \State $\tilde{X} = \text{Augment}(\text{Normalize}(\text{Filter}(X)))$
        \State $\{F_1, \dots, F_4\} = \text{FeatureExtractor}(\tilde{X}, \theta)$
        \State $H_{fusion} = \text{Concatenate}(\text{CP}_1(F_1), \dots, \text{CP}_4(F_4))$
        \State $\hat{Y} = \text{MLP}(H_{fusion})$
        \State Compute $\mathcal{L} = \text{CrossEntropy}(\hat{Y}, Y)$ with label smoothing
        \State Update $\theta$ via $\nabla_{\theta}\mathcal{L}$ and adaptive momentum
    \EndFor
    
    \State \textbf{Validation Phase:} Calculate $F1_{curr}$ on $\mathcal{L}_{vl}$.
    \State Step $\mathcal{S}$ and update $\theta_{best}$ if $F1_{curr} > F1_{best}$.
    
    \If{patience limit $P$ is reached}
        \State \textbf{break} 
    \EndIf
\EndFor
\State \Return $\theta_{best}$
\end{algorithmic}
\end{algorithm}

\subsection{Dataset Overview}

To critically analyze the cross-domain robustness of the HeartBeatAI framework, we utilized a unified cohort of 42,555 samples from four large-scale, heterogeneous public ECG datasets: CPSC2018 \cite{liu2018_cpsc}, PTB-XL \cite{wagner2020}, Georgia \cite{alday2020_challenge}, and Chapman \cite{zheng2020_chapman}. These datasets were selected to capture realistic clinical variations in patient demographics, recording devices, and signal lengths (ranging from 6 to 60 seconds, all standardized to a 500 Hz sampling rate). 

This study focuses on the classification of seven broad categories: Normal (N), Atrial Fibrillation (AF), Premature Atrial Contraction (PAC), Premature Ventricular Contraction (PVC), Left Bundle Branch Block (LBBB), Right Bundle Branch Block (RBBB), and First-degree Atrioventricular Block (I-AVB). Concentrating on these clinically dominant arrhythmias ensures sufficient sample sizes with clear morphological features, mitigating statistical noise from rare outlier pathologies across global hospital systems \cite{who2021_cvd, thygesen2018_udmi}. Detailed characteristics, dataset-specific sample sizes, and label distributions are comprehensively outlined in Table \ref{tab:demographics}.

The aggregated cohort presents a rigorous test for domain generalization across diverse populations spanning three continents. Excluding PTB-XL (which lacks specific demographic metadata but remains vital for morphological feature learning) \cite{wagner2020}, the mean patient age across the remaining sources is $60.2 \pm 17.2$ years, with a gender distribution of 54.8\% male and 45.2\% female. This diverse, balanced representation minimizes demographic bias and reinforces the clinical legitimacy of our LODO evaluation \cite{zhou2022_survey}.

\begin{table*}[htbp]
\centering
\caption{Comprehensive overview of dataset demographics and arrhythmia class distribution across the unified cohort ($N=42,555$).}
\label{tab:demographics}
\resizebox{\textwidth}{!}{%
\begin{tabular}{lccccccccccccc}
\toprule
\multirow{2}{*}{\textbf{Dataset}} & \multicolumn{4}{c}{\textbf{Demographics}} & \multicolumn{8}{c}{\textbf{Arrhythmia Class Distribution}}\\
\cmidrule(lr){2-5} \cmidrule(lr){6-13}
 & Subjects ($n$) & Age (Yrs) & Male (\%) & Female (\%) & AF & I-AVB & LBBB & N & PAC & PVC & RBBB & Total Samples \\
\midrule
CPSC2018 \cite{liu2018_cpsc} & 6,250 & 60.5$\pm$19.1 & 54.1\% & 45.9\% & 1,221 & 722 & 199 & 918 & 544 & 971 & 1,675 & 6,250 \\
Chapman \cite{zheng2020_chapman} & 9,501 & 60.3$\pm$16.8 & 56.6\% & 43.4\% & 2,224 & 0 & 0 & 7,277 & 0 & 0 & 0 & 9,501 \\
Georgia \cite{alday2020_challenge} & 6,676 & 59.9$\pm$15.8 & 52.8\% & 47.2\% & 258 & 343 & 610 & 4,925 & 99 & 17 & 424 & 6,676 \\
PTB-XL \cite{wagner2020} & 18,885 & N/A & N/A & N/A & 1,307 & 566 & 414 & 15,950 & 341 & 1,143 & 407 & 20,128 \\
\midrule
\textbf{Total} & \textbf{41,312*} & \textbf{60.2$\pm$17.2} & \textbf{54.8\%} & \textbf{45.2\%} & \textbf{5,010} & \textbf{1,631} & \textbf{1,223} & \textbf{29,070} & \textbf{984} & \textbf{2,131} & \textbf{2,506} & \textbf{42,555} \\
\bottomrule
\end{tabular}%
}
\vspace{1mm}
\begin{flushleft}
\footnotesize \textit{*Note: Demographic average Age (Mean $\pm$ Std), Gender are calculated based on CPSC2018, Chapman, and Georgia datasets. Total subject count includes the distinct patients from PTB-XL.}
\end{flushleft}
\end{table*}

\subsection{Evaluation Protocols}
To ensure a robust and comprehensive evaluation of the HeartBeatAI framework, two distinct data partitioning protocols are implemented.

\subsubsection{Intra-Source Evaluation}
This evaluation determines the theoretical learning capacity of the framework. The entire dataset across the individual source domains (CPSC2018, PTB-XL, Georgia, and Chapman) is split into training, validation, and testing subsets, with the standard distribution of 70\%, 10\%, and 20\%, respectively \cite{strodthoff2021_jbhi}.

\subsubsection{LODO}
This setup is used to mimic a real-world clinical setting and to assess cross-domain generalization. In this case, each dataset is treated as the unseen target domain to assess zero-shot transfers to unseen, to evaluate the framework's ability to generalize to completely new data. The framework is developed using data from $N-1$ source domains and is tested with the remaining target domain data. This is done $N=4$ times, such that each individual dataset (CPSC2018, PTB-XL, Georgia, and Chapman) is used as the target domain once \cite{wang2022_survey}.

\subsection{Data Preprocessing}
To extract strong and dependable features from heterogeneous datasets, this study adopts a streamlined three-step preprocessing pipeline that enhances the signal-to-noise ratio (SNR) while strictly normalizing input dimensions \cite{li2021_generalization} as illustrated in Algorithm \ref{alg:training}.

\subsubsection{Digital Filtering and Artifact Removal}
Raw ECG signals are naturally impacted by power line interference, EMG noise, and baseline noise due to respiration and movement of the electrodes. To remove these artifacts, the signals undergo a second-order Butterworth bandpass filter with a passband from $0.5$ Hz to $45$ Hz. The lower cutoff of $0.5$ Hz will attenuate low-frequency baseline drift \cite{wang2020_adversarial}. The upper cutoff of $45$ Hz removes high-frequency noise while allowing the QRS Complex and P-wave morphology to remain clinically usable \cite{hong2020_review}.

\subsubsection{Statistical Amplitude Normalization}
Z-score normalization is performed on each lead independently that make the framework invariant to absolute voltage scales and is focused on relative morphological changes and the rhythm patterns obtained from different clinical settings \cite{ioffe2015_batchnorm}.

\subsubsection{Temporal Windowing and Segmentation}
As part of mini-batch processing for deep neural networks, continuous records of ECG are translated into representations as of a fixed size \cite{ribeiro2020}. Each signal is split into windows of $L = 5000$ samples. Assuming a sampling frequency of $500$ Hz, this means a window captures a $10$-second interval \cite{wagner2020}. This is long enough to capture complex arrhythmias, including Atrial Fibrillation (AF) and bundle branch blocks \cite{hannun2019_nature}.

For records where the length $len(x) < L$, we use zero-padding for the suffix to keep things uniform \cite{ballas2024_domain}. For longer records, we use center-cropping during the evaluation phase and random-offset cropping during the training phase. This type of cropping serves as stochastic temporal data augmentation, which is effective in improving the framework's generalization \cite{hong2020_review}.

\subsection{Structure of HeartBeatAI Framework}
\begin{figure*}[!htbp]
    \centering
    \includegraphics[width=0.75\textwidth]{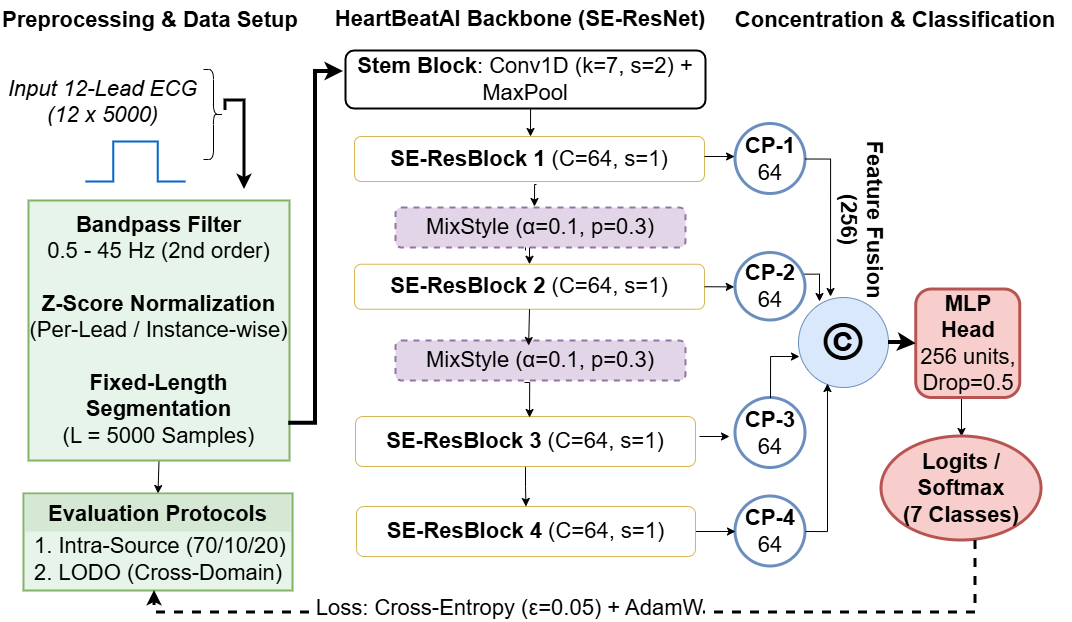} 
    \caption{Architectural diagram of the HeartBeatAI framework, highlighting the integration of the SE-ResNet backbone, MixStyle-driven domain generalization,
and the Multi-Layer Concentration Pipeline.}
    \label{fig:image5}
\end{figure*}

The proposed HeartBeatAI framework is designed to address domain shift failures by extracting domain-invariant features. Unlike traditional \enquote{black-box} frameworks, HeartBeatAI incorporates specific architectural mechanisms to force the learning of physiologically relevant features. The framework consists of two core components: the SE Backbone and the Multi-Layer Concentration Pipeline, as illustrated in Fig. \ref{fig:image5}.

\begin{figure}[!t]
    \centering
    \includegraphics[width=0.75\textwidth]{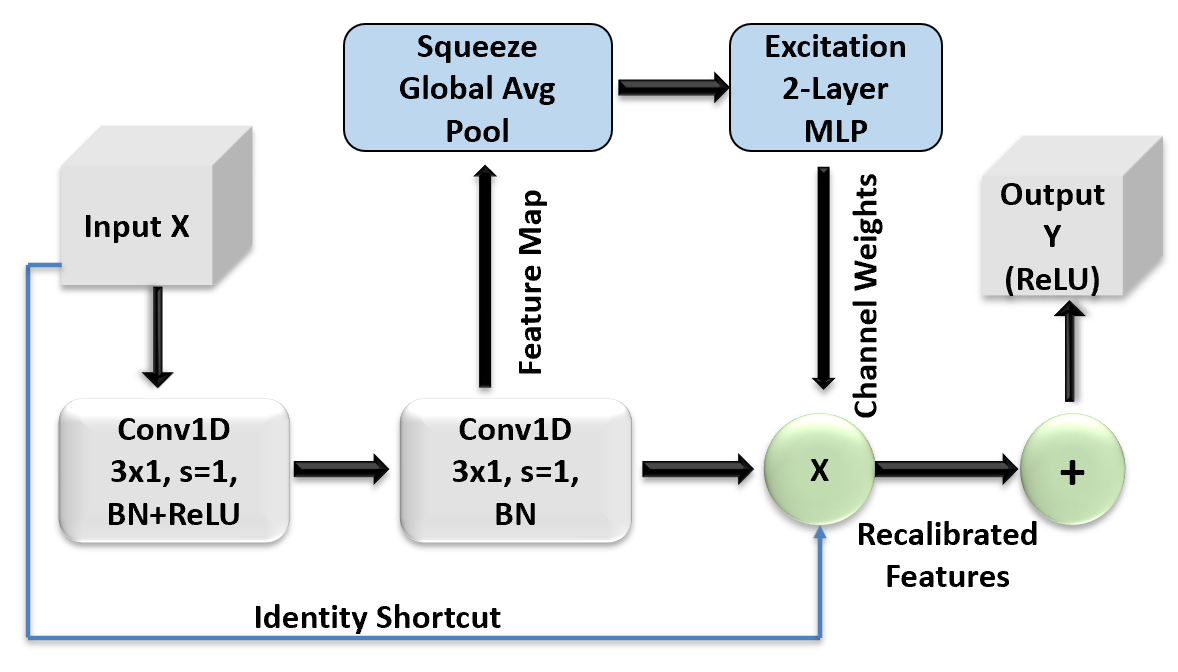} 
    \caption{Detailed architecture of the SE-ResBlock ($C=64, stride=1$), showing the \enquote{Squeeze} (Global Average Pooling) and \enquote{Excitation} (2-layer MLP) mechanisms for channel-wise feature recalibration.}
    \label{fig:seresblock}
\end{figure}

\subsubsection{SE Block}
Standard Convolutional Neural Networks (CNNs) treat all input channels with equal inductive bias. While appropriate for the RGB channels of standard images, this is fundamentally misaligned with 12-lead ECG analysis, where diagnostic information is heavily skewed depending on the pathology (e.g., Lead II for rhythm, precordial leads for bundle branch blocks). SE blocks are integrated to act as a simulated \enquote{diagnostic attention} mechanism. This dynamically recalibrates lead importance per instance by:
\begin{itemize}
    \item Elevating physiologically relevant leads based on pathology (e.g., Lead II for rhythm).
    \item Suppressing channels dominated by transient artifacts, such as baseline wander or electrode motion.
\end{itemize}


To enhance the network's sensitivity to informative ECG leads while suppressing noise, SE blocks are employed within the residual backbone \cite{he2016_resnet, hu2018_senet} as illustrated in Fig. \ref{fig:seresblock}.

\subsubsection{Concentration Pipeline (Feature Aggregation)}
While traditional visual tasks rely heavily on final network layers, cardiac arrhythmias manifest across wildly varying temporal scales. For example, Atrial Fibrillation presents as global RR-interval irregularities, whereas inverted T-waves are localized, high-frequency aberrations. To prevent discarding these fine-grained details, the Multi-Layer Concentration Pipeline aggregates intermediate features. This ensures the final feature vector ($H_{final}$) preserves both macro-rhythm statistics and micro-morphological geometry.
  
Unlike standard ResNets that utilize only the final feature map, the proposed framework aggregates features from multiple intermediate layers to capture multi-scale representations \cite{zhou2025}. Let $\mathbf{F}_l \in \mathbb{R}^{C_l \times T_l}$ denote the feature map from the $l$-th residual layer. The concentration operation $\Phi$ consists of a $1\times1$ convolution ($W_{1\times1}$) to reduce dimensionality, followed by global average pooling (GAP):
\begin{equation}
\mathbf{h}_l = \Phi(\mathbf{F}_l) = \text{GAP}(\text{Dropout}(\text{ReLU}(W_{1\times1} * \mathbf{F}_l)))
\end{equation}
The final representation $\mathbf{H}$ is obtained by concatenating the extracted features from all $N$ selected layers:
\begin{equation}
\mathbf{H} = [\mathbf{h}_1, \mathbf{h}_2, \dots, \mathbf{h}_N]
\end{equation}

\subsubsection{Domain Generalization via MixStyle Regularization}
Domain shift in computer vision typically involves variations in lighting, background, or camera angle. As noted in Section I, domain shift in clinical electrophysiology manifests as fundamentally different statistical noise profiles driven by institutional protocols and devices. To mitigate this, we employ MixStyle regularization \cite{zhou2021_mixstyle}, which is uniquely suited to smooth out these deterministic device signatures. By probabilistically interpolating the mean and variance of intermediate feature maps between random samples during training, MixStyle simulates novel \enquote{virtual} domains.

Let $\mathbf{x} \in \mathbb{R}^{B \times C \times L}$ be a feature map within the residual layers, where $B$, $C$, and $L$ denote the batch size, number of channels, and temporal length, respectively. The mean $\mu$ and standard deviation $\sigma$ are computed across the temporal dimension $L$. To synthesize a novel style, a mixing coefficient $\lambda$ is sampled from a Beta distribution:
\begin{equation}
\lambda \sim \text{Beta}(\alpha, \alpha), \quad \text{where } \alpha=0.1
\end{equation}

Given a spatially shuffled instance $\mathbf{x}_{perm}$ from the same batch, the mixed feature statistics $\gamma_{mix}$ and $\beta_{mix}$ are derived as follows:
\begin{align}
\gamma_{mix} &= \lambda \sigma(\mathbf{x}) + (1-\lambda) \sigma(\mathbf{x}_{perm}) \\
\beta_{mix} &= \lambda \mu(\mathbf{x}) + (1-\lambda) \mu(\mathbf{x}_{perm})
\end{align}

The stylized feature map $\hat{\mathbf{x}}$ is then reconstructed by applying the mixed statistics to the normalized input \cite{ioffe2015_batchnorm}:
\begin{equation}
\hat{\mathbf{x}} = \gamma_{mix} \left( \frac{\mathbf{x} - \mu(\mathbf{x})}{\sigma(\mathbf{x})} \right) + \beta_{mix}
\end{equation}

This operation is applied during training with a probability $p=0.3$. This stochastic perturbation forces the network to prioritize the domain-invariant physiological features discussed previously \cite{zhou2021_mixstyle}.

\subsubsection{Classification and Optimization}
\label{subsubsec:classification}

After feature aggregation from the Concentration Pipeline, the fused representation vector $H_{final} \in \mathbb{R}^{256}$ contains multi-scale morphological information, and then the vector is fed to the last classification head.

\paragraph{Multilayer Perceptron (MLP) Head}

The classifier is structured as a dense MLP to project the high-dimensional features into the relevant diagnostic classes. It has a single hidden linear layer with 256 neurons and a ReLU activation function. To address overfitting, which is a major concern on high-dimensional ECG data, a Dropout layer is used with $p=0.5$ prior to the last output layer \cite{srivastava2014_dropout}.

The last projection assigns the latent features to the class logits $z \in \mathbb{R}^{K}$ where $K=7$ is the number of target arrhythmia classes (N, AF, I-AVB, LBBB, RBBB, PAC, PVC). 

\paragraph{Objective Function with Label Smoothing}
Standard Cross-Entropy (CE) loss pushes the framework to predict the correct class with a probability of 1.0, which could result in overfitting and failure to generalize to unseen domains. The proposed framework uses Label Smoothing with a soft target distribution instead of a one-hot encoded target. This regularization technique penalizes over-confident predictions and clusters classes more tightly in the embedding space.

\subsection{Explainability Layer via 1D Grad-CAM (EX-AI)}
To move beyond the \enquote{black-box} DL models and clinical auditability, the HeartBeatAI framework uses an integrated explainability layer via 1D Grad-CAM \cite{jang2025}. This mechanism will help empirically trace the explainability of the diagnostic decision to particular morphological segments of the ECG, thereby making sure that the framework’s high performance is based on relevant physiological reasoning rather than dataset overfitting \cite{li2021_generalization}. The computation of the Grad-CAM saliency map is detailed in Algorithm \autoref{alg:gradcam}. Grad-CAM serves as “visual proof” for the framework’s internal reasoning, which are below: 
\begin{itemize} 

\item Physiological Alignment: The highlighted areas in the heatmap are in qualitative alignment with the QRS and T-wave complexes, hence it is clear that the network captures clinically relevant components for the detection of arrhythmias \cite{goettling2024}.

\item Lead-wise Validation: The saliency computation validates the SE blocks \cite{hu2018_senet} by illustrating how the framework recalibrates lead importance to concentrate on high-fidelity diagnostic components.

  \item Architectural Transparency: This layer, along with the Concentration Pipeline \cite{zhou2025}, allows visualization of the multi-scale representations so that the last 256-dimensional feature vector  $H_{final}$ contains the useful morphological changes.

\end{itemize}

\begin{algorithm}
\caption{1D Grad-CAM for Explainable Arrhythmia Detection}
\label{alg:gradcam}
\begin{algorithmic}[1]
\Require Trained Framework $M$, Target Convolutional Layer $L$, Input ECG Signal $X$, Target Class $c$
\Ensure Class-Discriminative Saliency Map $\mathcal{H}$
\State BEGIN
\State Extract feature map activations $A$ from layer $L$ during the forward pass of $M(X)$
\State Compute predicted class scores $\hat{y} = M(X)$
\State Isolate the unnormalized score for the target class: $S_c = \hat{y}_c$
\State Compute gradients of $S_c$ with respect to the layer activations $A$: $G = \frac{\partial S_c}{\partial A}$
\State \textbf{Channel Weighting:}
\For{each channel $k$}
    \State Calculate global average pooling of gradients over time dimension $T$: $w_k = \frac{1}{T} \sum_{t=1}^{T} G_{k,t}$
\EndFor
\State Compute the linear combination of activations weighted by $w_k$: $L_{Grad-CAM} = \text{ReLU} \left( \sum_k w_k A_k \right)$
\State \textbf{Normalization and Visualization:}
\State Normalize the heatmap to the range $[0, 1]$: $\mathcal{H} = \frac{L_{Grad-CAM}}{\max(L_{Grad-CAM})}$
\State Interpolate $\mathcal{H}$ to match the original temporal length of $X$
\State \Return $\mathcal{H}$
\State END
\end{algorithmic}
\end{algorithm}

\subsection{Implementation Details and Hyperparameter Configuration}
The HeartBeatAI framework is implemented in Python 3.8 with PyTorch \cite{paszke2019_pytorch}. All the experiments have been done using a workstation with an Intel Core i7 CPU, 32 GB RAM, and an NVIDIA GPU with CUDA for the heavy multi-domain ECG processing tasks. For our experiments, we chose the AdamW optimizer with a weight decay of $1 \times 10^{-4}$ \cite{loshchilov2019_adamw}, which decouples weight decay and gradient updates. This configuration improves generalization across ECG datasets, while regularization works under the stochastic shifts caused by MixStyle. Training begins with a learning rate of $1 \times 10^{-3}$ which is adjusted automatically through a \textit{ReduceLROnPlateau} scheduler, based on the validation Macro-F1 score. If there is no improvement, the scheduler decreases the learning rate by a factor of 0.5 after 5 epochs. To minimize the use of computational resources and reduce overfitting, Early stopping is implemented with a patience of 15 epochs and a maximum training limit of 50 epochs with a batch size of 64. This further increased the framework's robustness by using cross-entropy loss with label smoothing ($\epsilon = 0.05$), which helps in softening overconfident predictions and thus helps in generalization.








\nocite{*}
\bibliography{sn-bibliography}

\end{document}